\documentclass[10pt,twocolumn,letterpaper]{article}
\usepackage{wacv}
\usepackage{colortbl}
\definecolor{lowpbg}{HTML}{FDECEC}
\usepackage{booktabs}
\usepackage{array}
\usepackage{tabularx}
\newcolumntype{R}{>{\raggedleft\arraybackslash}X}
\usepackage{multirow}
\usepackage{placeins}
\usepackage{tikz}
\usetikzlibrary{positioning,arrows.meta,calc}
\definecolor{lowpred}{HTML}{D62728}
\definecolor{lowpteal}{HTML}{0F8B8D}
\definecolor{ink}{HTML}{1F2937}
\definecolor{slate}{HTML}{3B5B8C}
\definecolor{blockbg}{HTML}{F5F7FA}
\definecolor{blockedge}{HTML}{C9D2DE}
\definecolor{defdark}{HTML}{A61C1C}
\definecolor{deffill}{HTML}{FBE7E6}
\definecolor{defbar}{HTML}{F4C2C0}
\definecolor{glbdark}{HTML}{5B2C83}
\definecolor{glbfill}{HTML}{F0EAF6}
\definecolor{glbbar}{HTML}{DCCCEB}
\definecolor{glbmark}{HTML}{7030A0}
\definecolor{plandark}{HTML}{3D6B2F}
\definecolor{planfill}{HTML}{E4F0DC}
\definecolor{benchdark}{HTML}{7F6000}
\definecolor{benchfill}{HTML}{FFF4D1}
\definecolor{bbdark}{HTML}{595959}
\definecolor{bbfill}{HTML}{E7E6E6}
\definecolor{fireout}{HTML}{F26B1D}
\definecolor{firein}{HTML}{FFC233}
\definecolor{icecol}{HTML}{3D8FD1}

\newcommand{\keel}{LOWP}

\definecolor{wacvblue}{rgb}{0.21,0.49,0.74}
\usepackage[breaklinks,colorlinks,allcolors=wacvblue]{hyperref}
\usepackage[capitalize]{cleveref}
\graphicspath{{figs/}}

\def\wacvPaperID{1}
\def\confName{WACV}
\def\confYear{2027}

\title{Look Only Where You Plan for Real-Time End-to-End Urban Driving}

\author{Haohua Que\quad Tianle Zhu\quad Qianyi Wu\quad Zhipeng Bao\quad Handong Yao$^{*}$\\
University of Georgia}
\hypersetup{
  pdftitle={Look Only Where You Plan for Real-Time End-to-End Urban Driving},
  pdfauthor={Haohua Que, Tianle Zhu, Qianyi Wu, Zhipeng Bao, Handong Yao},
  pdfsubject={WACV 2027 UrbanState Workshop},
  pdfkeywords={end-to-end driving, motion planning, sparse queries, efficiency}
}

\begin{document}
\twocolumn[{%
\renewcommand\twocolumn[1][]{#1}%
\maketitle
\vspace{-1.6em}
\begin{center}
\includegraphics[width=\linewidth]{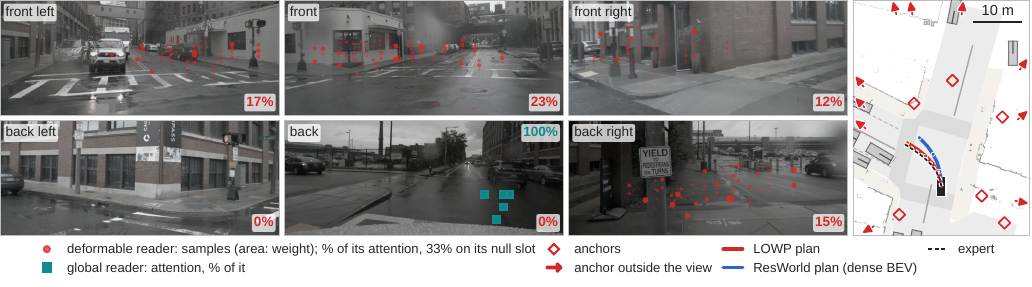}
\captionof{figure}{\keel{} reads a scene through 32 queries and builds no BEV (nuScenes, a left turn). Red: the deformable reader's last-layer sampling points (area: attention weight) and attention per camera; a third goes to its null slot. Teal: the global reader's last-layer attention, concentrated on a few rear-camera cells in the eight validation samples inspected. Right: the anchors (diamonds), the 3\,s plans of \keel{} and dense ResWorld, and the expert. Videos: \url{https://anonymous.4open.science/w/lowp-site-4598/}.}
\label{fig:teaser}
\end{center}
\vspace{0.4em}
}]
\begingroup
\renewcommand{\thefootnote}{\fnsymbol{footnote}}
\footnotetext[1]{Corresponding author: \href{mailto:Handong.Yao@uga.edu}{Handong.Yao@uga.edu}.}
\endgroup
\begin{abstract}
Dense bird's-eye-view (BEV) construction accounts for 60\% of the parameters in a strong recent driving planner.
This paper presents \keel{}, a planner that looks only where it plans: two sets of sixteen learnable queries read the cameras through the known calibration, one by sampling image features at projected 3D anchors and one by attending to features labeled with their viewing rays, so no dense BEV is built and no object or map is supervised.
Evaluations on open-loop nuScenes, pseudo-simulated NAVSIM and closed-loop Bench2Drive show competitive planning with a compact representation.
On nuScenes, average L2 changes from 0.302 to 0.296\,m with 53\% of the dense counterpart's parameters and 1.7 times its speed on the same GPU; the anchor-based reader performs better on turns than the global reader.
On NAVSIM, scoring the planner's own trajectories alongside a vocabulary gives 85.9 PDMS without retraining. The closed-loop control configuration scores 65.5 on 218 routes, with a brake-induced standstill failure whose release rule has an inconclusive aggregate benefit.
We distinguish measured latency from hardware-transferred estimates, which do not establish a common-hardware ranking.
\end{abstract}

\section{Introduction}
\label{sec:intro}

Urban autonomous mobility needs planners that react to dense traffic within the frame budget of on-vehicle compute.
End-to-end planners map sensor input directly to a future trajectory~\cite{uniad,vad,transfuser,chen2024e2esurvey}, and most first lift camera features into a dense bird's-eye-view grid~\cite{lss,bevdet,bevformer}.
That grid is built for perception: it covers every cell around the vehicle at a fixed resolution, whatever the trajectory depends on.
In a strong dense planner, ResWorld~\cite{resworld}, the depth network and BEV encoder hold 48.5\,M of its 80.8\,M parameters.

Three families of planners address this cost, and each keeps part of it.
Dense BEV planners such as UniAD~\cite{uniad}, VAD~\cite{vad} and ResWorld pay for the full reconstruction and add temporal modeling on top.
Sparse planners such as SparseDrive~\cite{sparsedrive}, DiffusionDrive~\cite{diffusiondrive}, MomAD~\cite{momad} and BridgeAD~\cite{bridgead} drop the grid but keep detection and mapping heads over hundreds of instances, with the annotations they require.
Vocabulary planners such as Hydra-MDP~\cite{hydramdp}, GTRS~\cite{gtrs}, DriveSuprim~\cite{drivesuprim} and iPad~\cite{ipad} rank thousands of candidate trajectories, and their best results rely on large backbones or several selection stages.
We investigate how little of the scene representation a planner needs to retain.

\keel{} reads only where it plans (\cref{fig:teaser}).
Two readers of sixteen queries replace the BEV: a deformable reader whose queries carry metric 3D anchors projected into the cameras through the known calibration, and a global reader whose queries attend to image cells labeled with the 3D points along their rays.
Each feeds its own planning head, and the driving command mixes the two trajectories.
No dense grid is built and no object or map is supervised (on nuScenes a depth adapter is trained with LiDAR depth, as in the dense counterpart).
Each regime adds a light head: regression in open loop, a vocabulary scorer that also ranks the planner's own trajectories in pseudo-simulation, and a control branch in closed loop.
The central result is a reduction in representation size and measured network latency while retaining competitive planning accuracy. Comparisons with external methods provide context, with hardware and controller differences stated explicitly.
Our contributions are:
\begin{enumerate}
\item \textbf{BEV-free calibrated query routing.} A scene representation of 32 calibrated tokens replaces the counterpart's 48.5\,M dense BEV pathway with 10.6\,M of sparse machinery, without object or map labels. It retains comparable nuScenes accuracy on two seeds at lower measured latency.
\item \textbf{Two experts mixed by the command.} The driving command mixes the readers' planning outputs. On turns, the anchor-based head has 7 collisions at 3\,s against 24 for the global head; mixing lowers L2 relative to either head in both driving regimes (\cref{tab:readers}).
\item \textbf{Evaluation across three regimes.} The shared design supports regression, candidate scoring and closed-loop control. Adding regressed candidates improves NAVSIM by 0.54 PDMS without retraining. Closed-loop comparisons document each controller and recovery rule, including a control-induced standstill failure and the tradeoffs of releasing its brake.
\end{enumerate}

\section{Related Work}
\label{sec:related}

\noindent\textbf{Dense end-to-end planners.}
ST-P3~\cite{stp3}, UniAD~\cite{uniad} and VAD~\cite{vad} build a BEV representation and stack perception, prediction and planning on it; PARA-Drive~\cite{paradrive} runs these modules in parallel, and GenAD~\cite{genad} and LAW~\cite{law} add generative or latent world modeling.
ResWorld~\cite{resworld} models temporal residuals of the BEV and is the dense counterpart \keel{} is built against: the two share the image backbone and the nuScenes protocol, and differ in the scene representation.
On nuScenes, ego status alone is a strong open-loop baseline~\cite{admlp,bevplanner}, which is why we report closed-loop and pseudo-simulated results next to open-loop L2.

\noindent\textbf{Sparse and query-based planners.}
Query-based 3D detection samples image features at projected reference points instead of building a grid~\cite{detr3d,petr,sparsebev,sparse4d}, using deformable attention~\cite{zhu2021deformable}.
SparseDrive~\cite{sparsedrive}, SparseAD~\cite{sparsead}, DiffusionDrive~\cite{diffusiondrive}, MomAD~\cite{momad}, BridgeAD~\cite{bridgead}, DiFSD~\cite{difsd} and FlowAD~\cite{flowad} carry this idea into planning but keep instance-level perception heads and their supervision.
HiP-AD~\cite{hipad} projects planning waypoints into the cameras and DriveTransformer~\cite{drivetransformer} attends from task queries to raw image features; both keep perception tasks.
\keel{} uses calibrated sampling for the planner alone: its queries are not supervised to be objects or map elements.

\noindent\textbf{Trajectory vocabularies, closed loop and speed reporting.}
Hydra-MDP~\cite{hydramdp} distills rule-based metrics into a scorer over a trajectory vocabulary; Hydra-MDP++~\cite{hydramdppp}, GTRS~\cite{gtrs}, DriveSuprim~\cite{drivesuprim}, iPad~\cite{ipad} and PRIX~\cite{prix} refine the candidates, the scorer or the backbone, and NAVSIM~\cite{navsim} evaluates them by non-reactive pseudo-simulation.
In CARLA~\cite{carla}, Bench2Drive~\cite{bench2drive} measures closed-loop driving over 220 short routes; TCP~\cite{wu2022tcp}, DriveAdapter~\cite{jia2023driveadapter}, ETA~\cite{eta} and Hydra-NeXt~\cite{hydranext} show that how the plan is turned into control matters as much as the plan. \keel{} adopts a control branch following Hydra-NeXt and analyzes its standstill failures and an experimental recovery rule.
Across these lines, frame rates are reported on different GPUs, at different precisions and for different parts of the pipeline, and many methods release no weights; we therefore transfer our own speed onto their hardware through anchors we can run, and drop an anchor whose reported speed is inconsistent with its network time.

\section{Method}
\label{sec:method}

\begin{figure*}[t]
\centering
\input{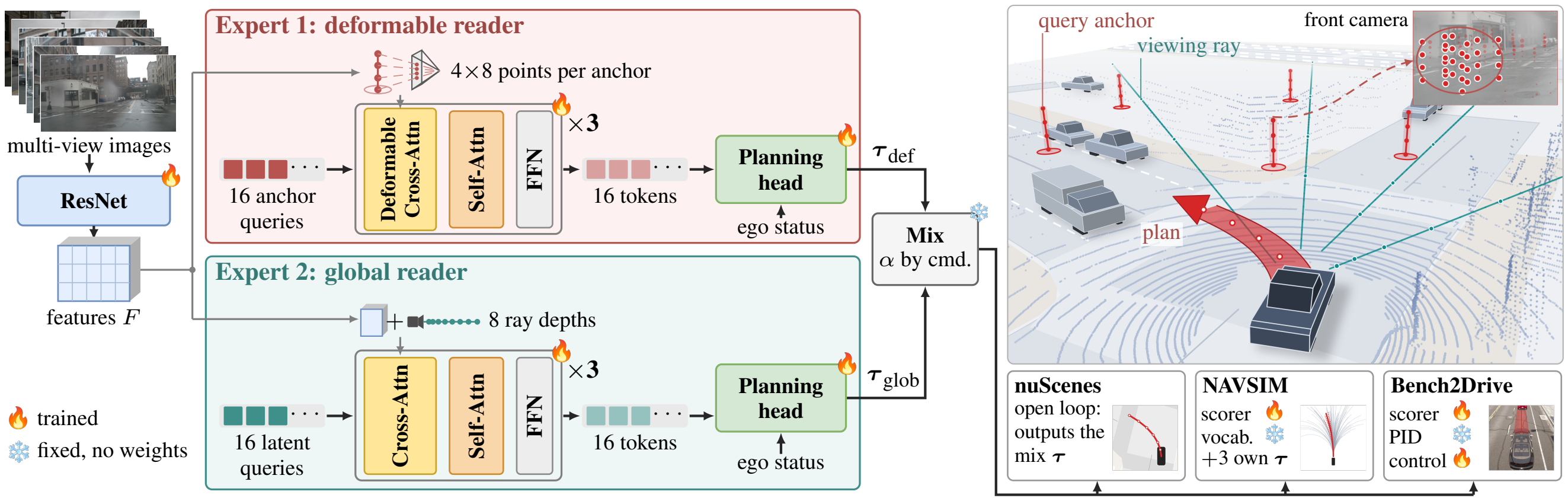}
\caption{\keel{}. Left: a ResNet trunk turns the camera images into stride-16 features $F$, and no dense BEV is built. Two experts plan from $F$, each a reader of 16 queries (three layers) and a planning head with the ego status. The deformable reader moves one metric anchor per query, lifts it to four heights of eight points, projects them through the calibration and aggregates the samples with a null slot; the global reader cross-attends to every cell of $F$, tagged with the encoding of its viewing ray. The command mixes the two trajectories, and a light head adapts the output to each benchmark. Right: both readers on a real nuScenes sample (red: the anchors, and where one anchor's 32 samples land in the front camera; teal: viewing rays, dotted at depths of the ray encoding, which spans 1 to 60\,m in eight steps); LiDAR points and vehicles are context only.}
\label{fig:overview}
\end{figure*}

\keel{} turns multi-camera images and the ego status into a future trajectory (\cref{fig:overview}). A ResNet~\cite{resnet} trunk (ResNet-34 on NAVSIM and Bench2Drive; on nuScenes the counterpart's backbone with the same initialization) produces stride-16 features, which two readers of $Q{=}16$ queries read through the camera calibration.

\subsection{Calibrated query routing}
\label{sec:readers}

\noindent\textbf{Geometry.}
A point $\mathbf{p}$ in the ego frame enters camera $n$ as $\mathbf{K}_n\mathbf{R}_n^{\top}(\mathbf{p}-\mathbf{t}_n)$, followed by the fixed crop-and-resize of the input pipeline.
When several cameras are stitched into one image, as on NAVSIM, a projected point is valid only inside its camera's band of the feature map.

\noindent\textbf{Deformable reader.}
Each query owns a metric anchor in the ego frame, initialized on a grid ahead of the vehicle and shifted per scene by the query content.
Every layer refines the anchor by a bounded step ($\tanh$ scaled to a few meters), lifts it to a vertical stack of four heights over three meters, and seeds eight sampling points per height on a ring that the query offsets.
Every (camera, height, point) candidate is projected, sampled bilinearly, and aggregated by multi-head attention; a learned null candidate joins the aggregation, so a query that no camera sees takes a learned state instead of an undefined one.
The anchor position is fed back as an embedding, the driving command routes the queries, and anchors that fall outside every camera at initialization are re-seeded.
On nuScenes a depth adapter, supervised with LiDAR depth during training only, as in the dense counterpart, scores each candidate by its predicted depth and adds that score to the aggregation weights; the camera-only NAVSIM and Bench2Drive models have neither the adapter nor any depth supervision.

\noindent\textbf{Global reader.}
Sixteen latent queries attend to every cell of the feature map.
Each cell is labeled with the 3D points that its viewing ray passes through at eight depths between 1 and 60\,m, encoded and added to the cell feature, so the attention is geometry-aware without any depth supervision.

\noindent\textbf{Planning heads and command mixing.}
Each reader's tokens, together with an ego-status token, feed a two-layer transformer head that regresses $T$ future poses; a reader and its head form one expert.
Each head receives its own $L_1$ loss on the expert trajectory, while the image trunk is shared and jointly updated. At inference their outputs are mixed as
$\boldsymbol{\tau} = \alpha\,\boldsymbol{\tau}_{\mathrm{def}} + (1-\alpha)\,\boldsymbol{\tau}_{\mathrm{glob}}$, with $\alpha{=}0.5$ when the command is to go straight and $\alpha{=}0.8$ when it is a turn.
The deformable head performs better on turns, the two are similar on straight driving, and mixing lowers L2 in both regimes (\cref{tab:readers}).
On nuScenes the model also reads one history frame.

\subsection{Heads for each regime}
\label{sec:regimes}

\noindent\textbf{Open loop.}
The mixed trajectory is the output.

\noindent\textbf{Pseudo-simulation: vocabulary scorer with own candidates.}
For NAVSIM we add the GTRS~\cite{gtrs} 8{,}192-entry trajectory vocabulary.
Each entry is embedded and cross-attends, over three layers, to the 32 reader tokens and the status token; one head per PDM sub-score and one imitation head are trained on the released simulated sub-scores of every training scene, and the entry with the best combined score is emitted~\cite{hydramdp,gtrs}.
The scorer's layers attend only from candidates to the scene, never between candidates, so any trajectory can be scored exactly as a vocabulary entry would be.
At inference we therefore add the three regressed trajectories (the mix and each head's) as candidates: the scorer is unchanged, and the plan is the best of 8{,}195 candidates instead of 8{,}192.

\noindent\textbf{Closed loop: control branch.}
For Bench2Drive the scorer ranks a 4{,}096-entry vocabulary and also attends to the global reader's keys pooled to 256 positions.
A PID controller tracks the selected trajectory, as in the CARLA baselines~\cite{wu2022tcp}.
Following Hydra-NeXt~\cite{hydranext}, a control branch on the reader tokens predicts throttle, steering and a brake logit; its throttle and steering are averaged with the PID's, and its brake may add braking the PID did not ask for.
The reported control configuration also uses a recovery rule outside the learned model: after 3\,s under throttle without motion the car reverses for 1\,s, at most three times per route. The earlier PID-only run without this rule is reported separately in \cref{tab:control_protocol}. This reverse rule is distinct from the experimental brake release in \cref{sec:closedloop}; published Zoo scores do not share our recovery wrapper.

\subsection{Implementation for latency}
\label{sec:latency}

A planner of this size is bound by kernel launches at batch one, so every host synchronization in the forward pass costs time.
Three patterns caused them: reading configuration constants back from the GPU to slice the stitched map, rebuilding normalization tensors from Python lists in every layer, and converting diagnostics to Python floats.
Computing the slices once as integers, building constant tensors once per shape and keeping diagnostics on the device removes them without changing any arithmetic: all outputs are bit-identical.

\section{Experiments}
\label{sec:experiments}

\subsection{Benchmarks and protocols}

\noindent\textbf{Protocols.}
On nuScenes~\cite{nuscenes} we follow ResWorld's~\cite{resworld} validation protocol with ego status: L2 and collision rate at 1, 2 and 3\,s under endpoint (MAX) and temporal-average (AVG) conventions. The official checkpoint reproduces its paper (MAX L2 0.589 versus 0.59); paired intervals resample 150 scene clusters 2{,}000 times. The flagship trains 12 epochs at batch 8 on one H100 and uses epoch-6 EMA weights. This epoch was selected on validation data; selecting on half the scenes and scoring on the other half picks it in 493 of 500 random splits. The second seed uses epoch 6 without selection.
On NAVSIM~\cite{navsim}, we report PDMS on navtest (12{,}146 scenarios). The camera-only regression model trains 100 epochs with the devkit recipe; its warm-started scorer trains 30 epochs on the released GTRS vocabulary sub-scores~\cite{gtrs}.
On Bench2Drive~\cite{bench2drive} we use the base training split, 220-route suite, leaderboard timeouts and one seed, merging recorded routes with the official script. The four configurations yield 218, 217, 218 and 216 scored routes (\cref{tab:control_protocol}). Missing records differ across configurations; paired analyses use their common routes. External scores are published results with different recovery wrappers. A partial TCP-traj rerun and selected qualitative traces are not a complete matched-harness baseline.

\noindent\textbf{Speed.}
\Cref{tab:timing} separates directly measured network latency from published speeds. Local timings use GPU-resident inputs, eager FP32 and batch one, excluding preprocessing, transfer, control and simulation. Recorded summaries expose minimum and median latency and protocol differences between sessions.
For contextual plots, anchor $a$ gives the estimate $\widehat{F}_{\keel{},a}=F_a^{\mathrm{pub}}\,t_a^{\mathrm{local}}/t_{\keel{}}^{\mathrm{local}}$, using the same session and statistic for both times. This assumes that hardware and implementation scaling transfers across architectures; it is not a direct measurement on the published GPU.
For the NAVSIM model with its own candidates, October 4 and 18 records imply about 58--60\,FPS through DiffusionDrive but 36--38 through iPad's Bench2Drive configuration, both reported on an RTX 4090. Their disagreement limits cross-method ordering. TransFuser also implies a different scaling factor; this alone does not identify its cause. PRIX uses FP16 on an RTX 3090. FlowAD's speed is unreported; its architectural ancestry supplies no bound. TCP's control-network runtime and iPad's Bench2Drive runtime are not assigned to their trajectory-only and NAVSIM accuracy rows, respectively.

% Recorded summaries, not newly run experiments. See evidence/timing_manifest.json.
\begin{table}[t]
\centering
\footnotesize
\setlength{\tabcolsep}{3.2pt}
\newcommand{\gh}[1]{\multicolumn{6}{@{}l}{\textit{#1}}\\}
\newcommand{\fp}{\rlap{$^{\dagger}$}}
\begin{tabularx}{\linewidth}{@{}Xrrrr>{\hspace{4.5pt}}r@{}}
\toprule
& \multicolumn{2}{c}{Oct.\ 2} & \multicolumn{2}{c}{Oct.\ 18} & \\
\cmidrule(lr){2-3}\cmidrule(lr){4-5}
Model & min & med. & min & med. & Pub. \\
\midrule
\gh{nuScenes}
ResWorld & 41.35 & 57.22 & 31.29 & 32.28\fp & 15.4\,FPS$^{\ast}$ \\
\rowcolor{lowpbg} \keel{} & 24.42 & 37.67 & 17.55 & 17.75\fp & \\
\rowcolor{lowpbg} \keel{}-S (global only) & 16.02 & 22.15 & -- & -- & \\
\addlinespace
\gh{NAVSIM}
DiffusionDrive & 12.96 & 20.75 & 9.19 & 9.26 & 45\,FPS \\
TransFuser & 5.07 & 5.91 & 5.12 & 5.28 & 60\,FPS \\
\rowcolor{lowpbg} \keel{} regression & 4.93 & 6.44 & 5.43 & 5.51 & \\
\rowcolor{lowpbg} \keel{} scorer & 6.09 & 7.79 & 6.60 & 6.69 & \\
\rowcolor{lowpbg} \keel{} scorer + own cand. & -- & -- & 7.05 & 7.14 & \\
\addlinespace
\gh{Bench2Drive}
iPad & 20.00 & 36.46 & 11.15 & 11.22 & 43\,ms \\
TCP-traj & -- & -- & 2.32 & 2.34\fp & -- \\
\rowcolor{lowpbg} \keel{} control & 6.03 & 8.41 & 6.47 & 6.53\fp & \\
\bottomrule
\end{tabularx}
\caption{\textbf{Direct network latency in milliseconds} (minimum and median) on an RTX PRO 6000 Blackwell, eager FP32, batch one, using CUDA events and GPU-resident inputs. Preprocessing, data transfer, PID/recovery and simulator time are excluded. The anchor script uses three interleaved rounds of 50 warm-up and 300 timed forwards; medians are over round medians. $^{\dagger}$The footprint script uses 20/100 forwards per round and one nuScenes sample instead of 32, so sessions are not a controlled timing ablation. Other inputs are synthetic; nuScenes uses trained checkpoints, and the footprint run also loads Bench2Drive checkpoints. The Oct.\ 18 log records PyTorch 2.11.0+cu128 and an idle start. Pub.: published anchor speeds on an RTX 4090 ($^{\ast}$H100), separate from the local measurements; their precision, batch and exact timing scope remain unverified.}
\label{tab:timing}
\end{table}

\subsection{Accuracy against speed}

\begin{figure*}[t]
\centering
\includegraphics[width=\linewidth]{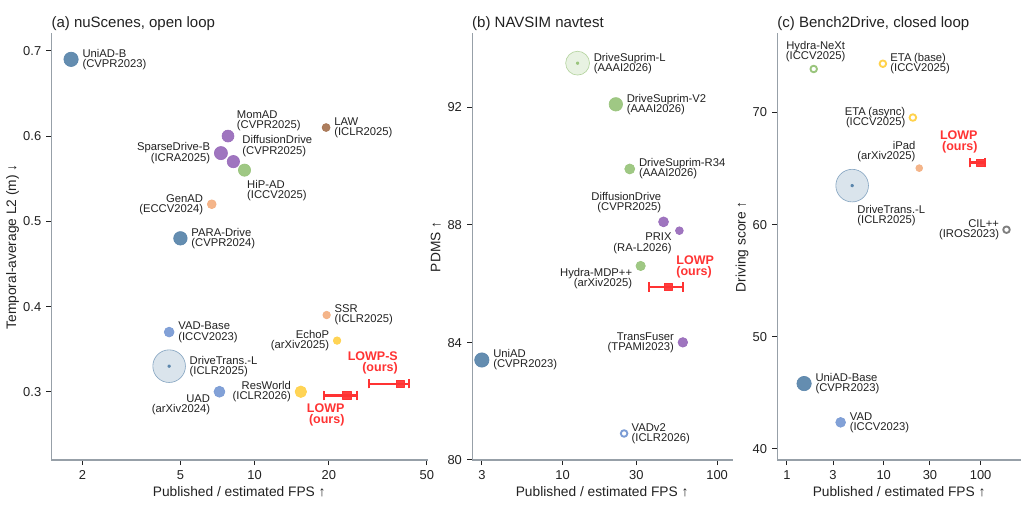}
\caption{Descriptive accuracy--speed comparison on (a) nuScenes (average L2, lower is better), (b) NAVSIM (PDMS) and (c) Bench2Drive (driving score). Red squares and bars are \keel{}'s transferred estimates and sensitivity ranges; circles use competitors' published speeds on heterogeneous hardware (area: parameters; open ring: not stated). These are not common-hardware Pareto fronts. Methods without a configuration-matched speed, including FlowAD, remain in the accuracy tables. \keel{}-S uses the global reader alone.}
\label{fig:fronts}
\end{figure*}

% generated by make_big_table.py; do not edit by hand
\begin{table*}[t]
\centering
\scriptsize
\setlength{\tabcolsep}{1.5pt}
\renewcommand{\arraystretch}{0.95}
\begin{tabular*}{\textwidth}{@{\extracolsep{\fill}}lrrr@{\hspace{5pt}}lrrr@{\hspace{5pt}}lrrr@{}}
\toprule
\multicolumn{4}{c}{\textbf{nuScenes} (open loop)} & \multicolumn{4}{c}{\textbf{NAVSIM} (pseudo-simulation)} & \multicolumn{4}{c}{\textbf{Bench2Drive} (closed loop)} \\
\cmidrule(lr){1-4}\cmidrule(lr){5-8}\cmidrule(lr){9-12}
Method & FPS$\uparrow$ & L2$\downarrow$ & Par. & Method & FPS$\uparrow$ & PDMS$\uparrow$ & Par. & Method & FPS$\uparrow$ & DS$\uparrow$ & Par. \\
\midrule
\cellcolor{lowpbg}\textbf{LOWP-S (global)} & \cellcolor{lowpbg}\textbf{29--42$^{E}$} & \cellcolor{lowpbg}\textbf{0.309} & \cellcolor{lowpbg}\textbf{35} & \cellcolor{lowpbg}\textbf{LOWP regr. (ours)} & \cellcolor{lowpbg}\textbf{47--86$^{E}$} & \cellcolor{lowpbg}\textbf{85.18} & \cellcolor{lowpbg}\textbf{31} & AD-MLP & 333 & 18.05 & 0 \\
\cellcolor{lowpbg}\textbf{LOWP (ours)} & \cellcolor{lowpbg}\textbf{19--26$^{E}$} & \cellcolor{lowpbg}\textbf{0.296} & \cellcolor{lowpbg}\textbf{43} & TransFuser & 60 & 84.00 & 56 & CIL++ & 185 & 59.53 & - \\
EchoP & 22 & 0.360 & 35$^{e}$ & PRIX & 57 & 87.80 & 37 & \cellcolor{lowpbg}\textbf{LOWP (ours)} & \cellcolor{lowpbg}\textbf{77--111$^{E}$} & \cellcolor{lowpbg}\textbf{65.51} & \cellcolor{lowpbg}\textbf{34} \\
ResWorld & 15 & 0.300 & 81 & \cellcolor{lowpbg}\textbf{LOWP (ours)} & \cellcolor{lowpbg}\textbf{36--60$^{E}$} & \cellcolor{lowpbg}\textbf{85.89} & \cellcolor{lowpbg}\textbf{34} & iPad & 23 & 65.02 & 29 \\
HiP-AD & 9.1 & 0.560 & 90 & DiffusionDrive & 45 & 88.10 & 60 & ETA (async) & 20 & 69.53 & - \\
DiffusionDrive & 8.2 & 0.570 & 92$^{e}$ & Hydra-MDP++ & 32 & 86.60 & 53 & ETA (base) & 9.8 & \textbf{74.33} & - \\
MomAD & 7.8 & 0.600 & 86$^{e}$ & DriveSuprim-R34 & 27 & 89.90 & 61 & DriveTrans.-L & 4.7 & 63.46 & 646 \\
SparseDrive-B & 7.3 & 0.580 & 105 & VADv2 & 25 & 80.90 & - & VAD & 3.6 & 42.35 & 58$^{e}$ \\
VAD-Base & 4.5 & 0.370 & 58$^{e}$ & DriveSuprim-V2 & 22 & 92.10 & 110 & Hydra-NeXt & 1.9 & 73.86 & - \\
DriveTrans.-L & 4.5 & 0.330 & 646 & DriveSuprim-L & 12 & \textbf{93.50} & 345 & UniAD-Base & 1.5 & 45.81 & 125 \\
UniAD-Base & 1.8 & 0.690 & 125 & UniAD & 3.0 & 83.40 & 125 & DriveAdapter & 1.1 & 64.22 & - \\
FlowAD & -- & \textbf{0.250} & 101 & iPad & -- & 91.70 & 29 & TCP-traj & -- & 59.90 & 27 \\
\bottomrule
\end{tabular*}
\caption{Descriptive comparison with published baselines; further rows appear in \cref{tab:nus,tab:nav,tab:b2d}. Unmarked FPS: published on heterogeneous hardware; $^{E}$: transferred estimates for \keel{} (shaded), spanning anchor/statistic choices (and two sessions for NAVSIM), not confidence intervals. Direct measurements are in \cref{tab:timing}. FlowAD, NAVSIM iPad and TCP-traj have no verified configuration-matched speed here (--). These speeds and the unmatched closed-loop protocols do not establish a global Pareto ordering. Par.: parameters in millions ($^{e}$ estimated); bold: best listed accuracy.}
\label{tab:big}
\end{table*}

\Cref{fig:fronts,tab:big} contextualize accuracy and model size; \cref{tab:timing} supports the measured efficiency comparison. On nuScenes, \keel{} uses 42.9\,M parameters versus ResWorld's 80.8\,M and takes 24.42 versus 41.35\,ms in the original session. The independent footprint session also measures lower latency but uses a different input-reuse protocol. On NAVSIM, the October 18 candidate model takes 7.05\,ms versus DiffusionDrive's 9.19\,ms and TransFuser's 5.12\,ms, with PDMS 85.9, 88.1 and 84.0, respectively. On Bench2Drive the original timing ratio to iPad's configuration is 3.3, while TCP-traj is faster in the footprint measurement. These network comparisons do not standardize published driving scores or establish a universal speed ranking.

\subsection{Open-loop evaluation: nuScenes and NAVSIM}

\begin{figure*}[t]
\centering
\includegraphics[width=\linewidth]{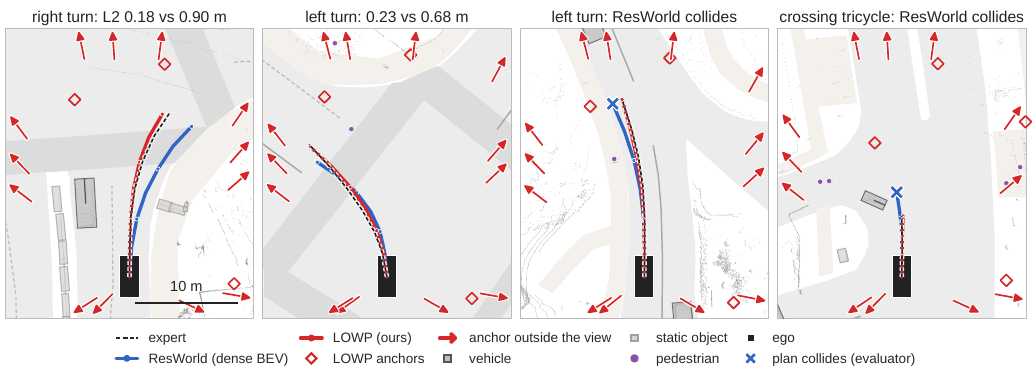}
\caption{nuScenes against dense ResWorld on four samples where plans differ (titles: average L2 of \keel{} and ResWorld, or an evaluator-flagged collision, marked by a cross). Bird's-eye view shows LiDAR (context only), map, annotated agents, \keel{}'s anchors (diamonds; arrows point to off-window anchors) and 3\,s plans.}
\label{fig:qual_nus}
\end{figure*}

\begin{figure*}[t]
\centering
\includegraphics[width=\linewidth]{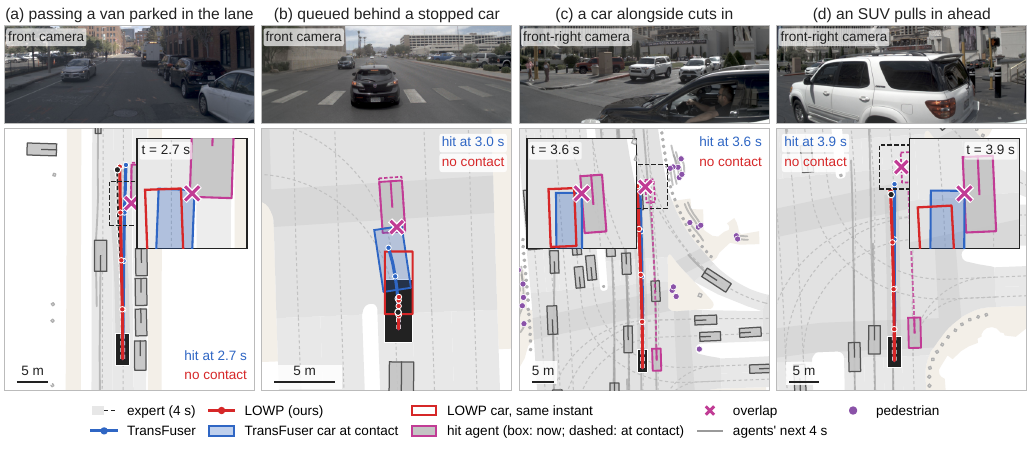}
\caption{NAVSIM against the official TransFuser baseline (retrained), on four selected scenes where TransFuser scores zero. Top: the camera showing the agent that is hit. Bottom: current agents and their next 4\,s (grey; hit agent dashed magenta), expert future (corridor and dashed line) and plans. Insets magnify first contact (cross); footprints follow the plans without the simulator's tracking controller.}
\label{fig:qual_nav}
\end{figure*}

% numbers from KEEL-ResWorld/keel_tools/reader_regime_table.py (per-sample evaluations, 5119 valid samples)
\begin{table}[tb]
\centering
\scriptsize
\setlength{\tabcolsep}{1.6pt}
\renewcommand{\arraystretch}{0.93}
\begin{tabularx}{\columnwidth}{@{}lRRRRRR@{}}
\toprule
& \multicolumn{3}{c}{Straight (4{,}436)} & \multicolumn{3}{c}{Turn (683)} \\
\cmidrule(lr){2-4}\cmidrule(lr){5-7}
Reading & AVG & MAX & Col. & AVG & MAX & Col. \\
\midrule
Deformable reader alone & 0.293 & 0.576 & 15 & 0.414 & 0.806 & \textbf{7} \\
Global reader alone & 0.291 & 0.585 & 13 & 0.438 & 0.909 & 24 \\
\rowcolor{lowpbg} Mix (\keel{}) & \textbf{0.280} & \textbf{0.553} & \textbf{11} & 0.401 & 0.784 & \textbf{7} \\
\keel{}-S (global model) & 0.290 & 0.578 & 13 & 0.435 & 0.894 & 18 \\
ResWorld (dense) & 0.287 & 0.562 & 14 & \textbf{0.397} & \textbf{0.760} & 8 \\
\bottomrule
\end{tabularx}
\caption{Which reader carries which regime on nuScenes: the two heads of one trained \keel{} alone and mixed, the separately trained global-reader model and the dense ResWorld, split by the driving command. L2 (m) under both conventions; Col.: collision events at 3\,s. On turns the deformable head beats the global head by 0.10\,m MAX L2 (95\% scene-bootstrap interval $[0.05, 0.16]$; AVG $[0.005, 0.047]$); on straight driving the two are level.}
\label{tab:readers}
\end{table}

\begin{table}[tb]
\centering
\scriptsize
\setlength{\tabcolsep}{2.2pt}
\begin{tabular*}{\columnwidth}{@{\extracolsep{\fill}}llcc@{}}
\toprule
Component & Metric & Before $\rightarrow$ after & Gain \\
\midrule
Two experts & nuS. L2 & 0.309 $\rightarrow$ 0.296 & $-0.013$ \\
Vocab. scorer & PDMS & 85.18 $\rightarrow$ 85.35 & $+0.17$ \\
Own candidates & PDMS & 85.35 $\rightarrow$ 85.89 & $+0.54$ $[0.35, 0.74]$ \\
Re-timed cand. & PDMS & 85.89 $\rightarrow$ 85.90 & $+0.01$ $[-0.03, 0.05]$ \\
Control config. & B2D DS & 60.56 $\rightarrow$ 65.72 & $+5.16 \pm 2.37$ \\
No host syncs & time (ms) & 9.44 $\rightarrow$ 6.09 & same output \\
\bottomrule
\end{tabular*}
\caption{Configuration comparisons (nuS.: nuScenes; B2D: Bench2Drive). Two experts compares separately trained \keel{}-S with the two-reader model, not a matched-capacity ablation. NAVSIM gains include 95\% scene-bootstrap intervals. B2D uses 217 common routes and mean $\pm$ standard error; the control model also receives additional training (\cref{tab:control_protocol}). Host-sync timing is the NAVSIM scorer, with identical outputs.}
\label{tab:abl}
\end{table}
% generated by make_big_table.py; do not edit by hand
\begin{table}[tb]
\centering
\fontsize{6}{6.9}\selectfont
\setlength{\tabcolsep}{1.6pt}
\renewcommand{\arraystretch}{0.93}
\begin{tabularx}{\columnwidth}{@{}lRRRRRR@{}}
\toprule
Method & NC & DAC & TTC & Comf. & EP & PDMS \\
\midrule
Ego-status MLP {\tiny(NeurIPS2024)} & 93.0 & 77.3 & 83.6 & \textbf{100.0} & 62.8 & 65.60 \\
VADv2 {\tiny(ICLR2026)} & 97.2 & 89.1 & 91.6 & \textbf{100.0} & 76.0 & 80.90 \\
UniAD {\tiny(CVPR2023)} & 97.8 & 91.9 & 92.9 & \textbf{100.0} & 78.8 & 83.40 \\
LTF {\tiny(TPAMI2023)} & 97.4 & 92.8 & 92.4 & \textbf{100.0} & 79.0 & 83.80 \\
TransFuser$^{\dagger}$ {\tiny(TPAMI2023)} & 97.7 & 92.8 & 92.8 & \textbf{100.0} & 79.2 & 84.00 \\
Hydra-MDP$^{\dagger}$ {\tiny(arXiv2024)} & 98.3 & 96.0 & 94.6 & \textbf{100.0} & 78.7 & 86.50 \\
Hydra-MDP++ {\tiny(arXiv2025)} & 97.6 & 96.0 & 93.1 & \textbf{100.0} & 80.4 & 86.60 \\
PRIX {\tiny(RA-L2026)} & 98.1 & 96.3 & 94.1 & \textbf{100.0} & 82.3 & 87.80 \\
DiffusionDrive$^{\dagger}$ {\tiny(CVPR2025)} & 98.2 & 96.2 & 94.7 & \textbf{100.0} & 82.2 & 88.10 \\
DriveSuprim (R34) {\tiny(AAAI2026)} & 97.8 & 97.3 & 93.6 & \textbf{100.0} & 86.7 & 89.90 \\
Hydra-MDP (V2-99) {\tiny(arXiv2024)} & 98.4 & 97.8 & 93.9 & \textbf{100.0} & 86.5 & 90.30 \\
Hydra-MDP (ViT-L) {\tiny(arXiv2024)} & 98.4 & 97.7 & 94.5 & \textbf{100.0} & 85.0 & 89.90 \\
iPad {\tiny(arXiv2025)} & \textbf{98.6} & 98.3 & 94.9 & \textbf{100.0} & 88.0 & 91.70 \\
DriveSuprim (V2-99) {\tiny(AAAI2026)} & 98.0 & 98.2 & 94.2 & \textbf{100.0} & 90.0 & 92.10 \\
DriveSuprim (ViT-L) {\tiny(AAAI2026)} & \textbf{98.6} & \textbf{98.6} & \textbf{95.5} & \textbf{100.0} & \textbf{91.3} & \textbf{93.50} \\
\rowcolor{lowpbg} LOWP regression & 97.6 & 94.0 & 92.7 & \textbf{100.0} & 80.1 & 85.18 \\
\rowcolor{lowpbg} LOWP scorer & 96.7 & 96.0 & 89.5 & 99.9 & 81.8 & 85.35 \\
\rowcolor{lowpbg} LOWP scorer + own cand. & 96.7 & 96.4 & 90.0 & 99.9 & 82.1 & 85.89 \\
\bottomrule
\end{tabularx}
\caption{NAVSIM navtest sub-scores and PDMS. $^{\dagger}$ camera and LiDAR; every \keel{} row is camera-only. Shaded: \keel{}; bold: best per column.}
\label{tab:nav}
\end{table}

\noindent\textbf{nuScenes.}
In the same evaluator (\cref{tab:nus,fig:qual_nus}), \keel{} changes temporal-average L2 from ResWorld's 0.302 to 0.296\,m (paired interval of the improvement $[0.0007,0.0116]$) and endpoint L2 from 0.589 to 0.584\,m. Collision point estimates are lower (0.143 versus 0.169\% endpoint; 0.056 versus 0.060\% average), but paired intervals include zero improvement. A second seed yields 0.298\,m average L2 and 0.048\% average collision rate. The small accuracy difference accompanies substantial parameter and latency reductions. The separately trained global model \keel{}-S uses 35.5\,M parameters and 16.0\,ms at 0.309\,m; FlowAD~\cite{flowad} reports lower L2 (0.25\,m) with unknown speed.
\Cref{tab:readers} splits samples by driving command. The deformable head cuts turn MAX L2 by 0.10\,m and 3\,s collisions from 24 to 7 relative to the global head of the same model. Mixing lowers L2 relative to either head in both regimes, while turn collisions remain at 7. The dense planner retains a small turn-L2 edge (0.760 versus 0.784\,m MAX). These head-output tests are distinct from independently training a deformable-only model or a matched-capacity pair of identical readers.

\noindent\textbf{NAVSIM.}
Hydra-MDP++~\cite{hydramdppp}, PRIX~\cite{prix} and DiffusionDrive~\cite{diffusiondrive} score 0.7 to 2.2 PDMS above \keel{} at reported speeds of 32 to 57\,FPS (\cref{tab:big}); transfer sensitivity prevents a definitive ordering against \keel{}. Against DiffusionDrive, \keel{} has similar drivable-area and ego-progress scores but lower no-collision (96.7 versus 98.2) and time-to-collision (90.0 versus 94.7) scores (\cref{tab:nav}). These locate the performance gap without isolating the effect of camera-only input or missing box supervision.

\subsection{Closed-loop evaluation: Bench2Drive}
\label{sec:closedloop}

% Historical run summaries: docs/REASONING_LOG.md, sections 172, 179, 186, 199.
% These are different scored route sets; use their intersections for paired gains.
\begin{table}[tb]
\centering
\scriptsize
\setlength{\tabcolsep}{2.3pt}
\begin{tabular*}{\columnwidth}{@{\extracolsep{\fill}}lcccc@{}}
\toprule
Controller & Reverse & Release & Routes & DS \\
\midrule
PID & off & off & 218 & 59.01 \\
PID & on & off & 217 & 60.56 \\
PID + control & on & off & 218 & 65.51 \\
PID + control & on & on & 216 & 67.10 \\
\bottomrule
\end{tabular*}
\caption{Bench2Drive controller and recovery configurations, one seed. Reverse: after 3\,s under throttle without motion, reverse for 1\,s, at most three times per route. Release: after 20\,s stopped, temporarily use PID longitudinal control when the head brakes against a moving plan. Paired DS changes use common routes: reverse $+1.30$ (217); control configuration $+5.16\pm2.37$ (217); release $+1.59\pm2.06$ (216), mean $\pm$ standard error where recorded. The control configuration is additionally trained; the release run also changes simulator process placement. These comparisons do not isolate either change from run variability.}
\label{tab:control_protocol}
\end{table}

\begin{figure*}[t]
\centering
\includegraphics[width=\linewidth]{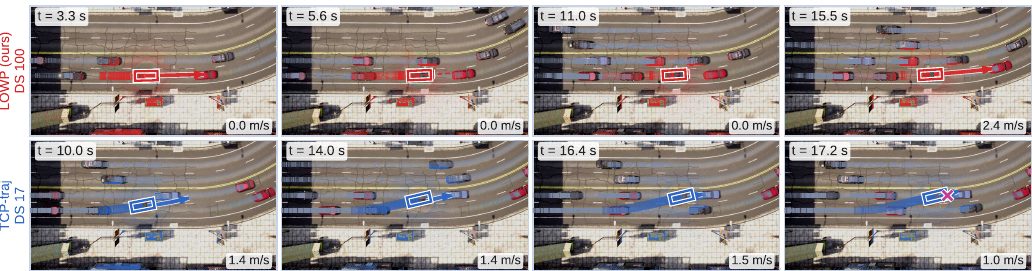}
\caption{Bench2Drive route 152 against TCP-traj (released weights in our harness), from selected traced runs, each row at its own instants. The ego is outlined; bands show driven paths and lines show plans (bars mark stationary plans). Simulated LiDAR is context only. Grey paths show other vehicles, and crosses mark contact. \keel{} stops about 5\,m behind a braking car and resumes when the queue moves; TCP-traj drifts left and rear-ends a stopped car (17.2\,s).}
\label{fig:qual_b2d}
\end{figure*}

% generated by make_big_table.py; do not edit by hand
\begin{table}[tb]
\centering
\fontsize{6}{6.9}\selectfont
\setlength{\tabcolsep}{1.2pt}
\renewcommand{\arraystretch}{0.93}
\begin{tabularx}{\columnwidth}{@{}lRRRRRRRR@{}}
\toprule
& \multicolumn{4}{c}{L2 (m)} & \multicolumn{4}{c}{CR (\%)} \\
\cmidrule(lr){2-5}\cmidrule(lr){6-9}
Method & 1s & 2s & 3s & Avg & 1s & 2s & 3s & Avg \\
\midrule
ST-P3 & 1.33 & 2.11 & 2.90 & 2.110 & 0.23 & 0.62 & 1.27 & 0.710 \\
UniAD-Base & 0.44 & 0.67 & 0.96 & 0.690 & 0.04 & 0.08 & 0.23 & 0.120 \\
VAD-Base & 0.17 & 0.34 & 0.60 & 0.370 & 0.07 & 0.10 & 0.24 & 0.140 \\
BEV-Planner & 0.16 & 0.32 & 0.57 & 0.350 & \textbf{0.00} & 0.29 & 0.73 & 0.340 \\
PARA-Drive & 0.25 & 0.46 & 0.74 & 0.480 & 0.14 & 0.23 & 0.39 & 0.250 \\
GenAD & 0.28 & 0.49 & 0.78 & 0.520 & 0.08 & 0.14 & 0.34 & 0.190 \\
LAW & 0.26 & 0.57 & 1.01 & 0.610 & 0.14 & 0.21 & 0.54 & 0.300 \\
LAW (+aux.) & 0.24 & 0.46 & 0.76 & 0.490 & 0.08 & 0.10 & 0.39 & 0.190 \\
SparseDrive-S & 0.29 & 0.58 & 0.96 & 0.610 & 0.01 & 0.05 & 0.18 & 0.080 \\
SparseDrive-B & 0.29 & 0.55 & 0.91 & 0.580 & 0.01 & \textbf{0.02} & 0.13 & 0.060 \\
Drive-OccWorld & 0.25 & 0.44 & 0.72 & 0.470 & 0.03 & 0.08 & 0.22 & 0.110 \\
SSR & 0.18 & 0.36 & 0.63 & 0.390 & 0.01 & 0.04 & 0.12 & 0.060 \\
MomAD & 0.31 & 0.57 & 0.91 & 0.600 & 0.01 & 0.05 & 0.22 & 0.090 \\
DiffusionDrive & 0.27 & 0.54 & 0.90 & 0.570 & 0.03 & 0.05 & 0.16 & 0.080 \\
BridgeAD-B & 0.28 & 0.55 & 0.92 & 0.580 & \textbf{0.00} & 0.04 & 0.20 & 0.080 \\
HiP-AD & 0.28 & 0.53 & 0.87 & 0.560 & 0.01 & 0.05 & 0.15 & 0.070 \\
DriveTransformer-L & 0.16 & 0.30 & 0.55 & 0.330 & 0.01 & 0.06 & 0.15 & 0.070 \\
EchoP & 0.17 & 0.33 & 0.58 & 0.360 & 0.03 & 0.05 & 0.14 & 0.070 \\
UAD & 0.13 & 0.28 & \textbf{0.48} & 0.300 & \textbf{0.00} & 0.12 & 0.55 & 0.220 \\
FocalAD & 0.27 & 0.57 & 0.96 & 0.600 & \textbf{0.00} & 0.04 & 0.24 & 0.090 \\
EgoFSD-B & \textbf{0.12} & 0.28 & 0.52 & 0.300 & \textbf{0.00} & 0.04 & 0.15 & 0.060 \\
ResWorld & 0.14 & 0.27 & 0.49 & 0.300 & 0.01 & 0.03 & 0.14 & 0.060 \\
FlowAD & - & - & - & \textbf{0.250} & - & - & - & \textbf{0.040} \\
\rowcolor{lowpbg} LOWP-S (global reader) & 0.14 & 0.28 & 0.51 & 0.309 & \textbf{0.00} & 0.03 & 0.17 & 0.067 \\
\rowcolor{lowpbg} LOWP (seed 1) & 0.14 & \textbf{0.27} & 0.48 & 0.296 & 0.02 & 0.03 & 0.11 & 0.056 \\
\rowcolor{lowpbg} LOWP (seed 3) & 0.14 & 0.27 & 0.49 & 0.298 & 0.01 & 0.02 & \textbf{0.11} & 0.048 \\
\bottomrule
\end{tabularx}
\caption{nuScenes, temporal-average protocol with ego status: L2 and collision rate (CR) at 1, 2 and 3\,s and their average. Shaded: \keel{}; bold: best per column. FlowAD reports only the averages.}
\label{tab:nus}
\end{table}

% generated by make_big_table.py; do not edit by hand
\begin{table}[tb]
\centering
\fontsize{6}{6.9}\selectfont
\setlength{\tabcolsep}{1.6pt}
\renewcommand{\arraystretch}{0.93}
\begin{tabularx}{\columnwidth}{@{}lRRRR@{}}
\toprule
Method & DS & SR (\%) & Effic. & Comf. \\
\midrule
AD-MLP {\tiny(arXiv2023)} & 18.05 & 0.00 & 48.45 & 22.63 \\
UniAD-Tiny {\tiny(CVPR2023)} & 40.73 & 13.18 & 123.92 & 47.04 \\
UniAD-Base {\tiny(CVPR2023)} & 45.81 & 16.36 & 129.21 & 43.58 \\
VAD {\tiny(ICCV2023)} & 42.35 & 15.00 & 157.94 & 46.01 \\
TCP {\tiny(NeurIPS2022)} & 40.70 & 15.00 & 54.26 & 47.80 \\
TCP-ctrl {\tiny(NeurIPS2022)} & 30.47 & 7.27 & 55.97 & \textbf{51.51} \\
TCP-traj {\tiny(NeurIPS2022)} & 59.90 & 30.00 & 76.54 & 18.08 \\
ThinkTwice {\tiny(CVPR2023)} & 62.44 & 31.23 & 69.33 & 16.22 \\
DriveAdapter {\tiny(ICCV2023)} & 64.22 & 33.08 & 70.22 & 16.01 \\
DriveTransformer-L {\tiny(ICLR2025)} & 63.46 & 35.01 & 100.64 & 20.78 \\
Hydra-MDP {\tiny(arXiv2024)} & 59.95 & 29.82 & 186.83 & 18.62 \\
CIL++ {\tiny(IROS2023)} & 59.53 & 25.57 & 204.23 & 18.38 \\
iPad {\tiny(arXiv2025)} & 65.02 & 35.91 & 161.31 & 28.21 \\
ETA (async) {\tiny(ICCV2025)} & 69.53 & 38.64 & 184.51 & 28.43 \\
ETA (base) {\tiny(ICCV2025)} & 74.33 & 48.33 & 186.04 & 25.77 \\
Hydra-NeXt$^{\ddagger}$ {\tiny(ICCV2025)} & 73.86 & 50.00 & 197.76 & 20.68 \\
DriveSuprim {\tiny(AAAI2026)} & 83.02 & 60.00 & \textbf{238.78} & 20.89 \\
HiP-AD {\tiny(ICCV2025)} & \textbf{86.77} & \textbf{69.09} & 203.12 & 19.36 \\
\rowcolor{lowpbg} LOWP & 60.56 & 30.88 & - & - \\
\rowcolor{lowpbg} LOWP + control branch & 65.51 & 34.40 & - & - \\
\midrule
\multicolumn{5}{l}{\textit{trained on PDM-Lite expert data (a larger training set)}} \\
TF++ {\tiny(arXiv2023)} & 84.21 & 67.27 & - & - \\
SimLingo {\tiny(CVPR2025)} & 85.94 & 66.82 & - & - \\
Hydra-NeXt {\tiny(ICCV2025)} & 86.00 & 68.18 & - & - \\
\bottomrule
\end{tabularx}
\caption{Bench2Drive, base split unless marked: driving score, success rate, efficiency and comfortness. External rows use their published protocols; shaded \keel{} rows use 217 recorded routes for PID and 218 with the control branch, both with standstill reversal (\cref{tab:control_protocol}). These are not matched-harness comparisons. $^{\ddagger}$ full 2\,M-frame training set. \keel{} efficiency and comfortness are unavailable (-); bold: best listed value per column.}
\label{tab:b2d}
\end{table}

The control configuration scores 65.51 with reverse recovery on and brake release off (\cref{tab:b2d,tab:control_protocol}). Published scores for iPad~\cite{ipad} (65.02), DriveAdapter~\cite{jia2023driveadapter} (64.22) and DriveTransformer-L~\cite{drivetransformer} (63.46) are numerically nearby; different wrappers, recorded routes and one seed prevent a claim of superiority or statistical equivalence. ETA~\cite{eta}, Hydra-NeXt~\cite{hydranext}, HiP-AD~\cite{hipad}, DriveSuprim~\cite{drivesuprim} and SimLingo~\cite{renz2025simlingo} report higher scores.
The existing PID-only runs without and with reverse yield 59.01 and 60.56 on their respective sets; on 217 common routes the gain is 1.30. This does not measure disabling reverse in the control model. Comparing the additionally trained control configuration with the PID/reverse baseline gives $+5.16\pm2.37$ DS on 217 common routes, a system-configuration comparison rather than a same-weight control-head toggle.

\noindent\textbf{Standstill deadlock.}
Vehicle collisions fall from 0.66 to 0.42 per route in the control configuration, but 26 of its 42 blocked routes end with the brake head holding still while the trajectory asks to move, mostly at 2 to 3.5\,m/s. This brake-held state does not trigger reverse recovery, which requires throttle.
The separate release run waits 20\,s at standstill, then uses PID longitudinal control when the head brakes against a moving plan, until speed reaches 2\,m/s or 3\,s pass. On 216 common routes it changes DS by $+1.59\pm2.06$, with no clear aggregate benefit: blocked routes fall from 41 to 24 while vehicle collisions rise from 0.44 to 0.62 per route. Simulator process/GPU placement also changes, so pairing does not isolate the rule from harness and run variability. Thirteen previously blocked routes complete without release firing. Brake release is absent from the reported 65.51 configuration.

\subsection{Qualitative results}

Videos are linked in \cref{fig:teaser}. In eight inspected nuScenes validation samples, deformable anchors move by at most a few meters and attention shares are similar across cameras. The global reader puts 94--100\% of its last-layer attention on five rear-camera cells and at most a fifth of its first-layer attention on the front camera. This localized observation does not establish which visual information causes its predictions or its contribution to the mixture.
The selected nuScenes cases (\cref{fig:qual_nus}) show \keel{} following turns where ResWorld under-turns or cuts a corner. In 24 extracted NAVSIM scenes where TransFuser scores zero and \keel{} does not, median sideways deviation from the expert is 0.8 versus 0.3\,m. \Cref{fig:qual_nav,fig:qual_b2d} illustrate agent interactions; these selected cases do not estimate benchmark-wide failure frequencies.

\subsection{Configuration analysis}

\Cref{tab:abl} separates observed changes in the evaluated configurations.
On NAVSIM, the vocabulary scorer alone adds little. Ranking the regressed trajectories alongside its vocabulary adds 0.54 PDMS with the same weights: 102 scenes leave a zero score and 57 enter one. Adding copies driven 0.8 to 1.2 times as fast changes PDMS by 0.008, with interval $[-0.030,0.051]$.
Removing the host synchronizations of \cref{sec:latency} preserves outputs and reduces the recorded NAVSIM-scorer latency from 9.44 to 6.09\,ms. Reported \keel{} times include this implementation work; anchor implementations are used as released and might also benefit from such changes.

\section{Conclusion and limitations}
\label{sec:conclusion}

\keel{} supports planning with 32 calibrated queries, without a dense BEV or object and map supervision. Its controlled nuScenes comparison shows comparable accuracy with 47\% fewer parameters and lower same-GPU latency. Candidate scoring and control extend the design to NAVSIM and Bench2Drive.

\noindent\textbf{Limitations.}
Transferred speeds do not establish a common-hardware Pareto ordering. Bench2Drive uses incomplete route sets and recovery wrappers differing from published baselines; its no-reverse comparison covers only the PID configuration. Paired route uncertainty does not measure seed variability. Matched-capacity, training-matched deformable-only and complete matched-wrapper controls remain absent. Head-output comparisons and eight attention visualizations do not isolate representation complementarity from capacity or ensembling. Only the nuScenes flagship has two seeds; other results use one.

\section*{Acknowledgments}
This work was supported in part by the Tribal and Rural Autonomous Vehicles for Efficiency, Livability and Safety (TRAVELS) project, funded by the U.S. Department of Transportation.
We thank the Georgia Advanced Computing Resource Center (GACRC), a partnership of the University of Georgia's research and information technology offices, for computational resources and technical assistance.

\FloatBarrier
{\small
\setlength{\bibsep}{0pt}
\bibliographystyle{ieeenat_fullname}
\bibliography{references}

\begin{thebibliography}{44}
\providecommand{\natexlab}[1]{#1}
\providecommand{\url}[1]{\texttt{#1}}
\expandafter\ifx\csname urlstyle\endcsname\relax
  \providecommand{\doi}[1]{doi: #1}\else
  \providecommand{\doi}{doi: \begingroup \urlstyle{rm}\Url}\fi

\bibitem[Caesar et~al.(2020)Caesar, Bankiti, Lang, Vora, Liong, Xu, Krishnan,
  Pan, Baldan, and Beijbom]{nuscenes}
Holger Caesar, Varun Bankiti, Alex~H. Lang, Sourabh Vora, Venice~Erin Liong,
  Qiang Xu, Anush Krishnan, Yu Pan, Giancarlo Baldan, and Oscar Beijbom.
\newblock {nuScenes}: A multimodal dataset for autonomous driving.
\newblock In \emph{Proceedings of the IEEE/CVF Conference on Computer Vision
  and Pattern Recognition (CVPR)}, 2020.

\bibitem[Chen et~al.(2024)Chen, Wu, Chitta, Jaeger, Geiger, and
  Li]{chen2024e2esurvey}
Li Chen, Penghao Wu, Kashyap Chitta, Bernhard Jaeger, Andreas Geiger, and
  Hongyang Li.
\newblock End-to-end autonomous driving: Challenges and frontiers.
\newblock \emph{IEEE Transactions on Pattern Analysis and Machine
  Intelligence}, 46\penalty0 (12), 2024.

\bibitem[Chitta et~al.(2023)Chitta, Prakash, Jaeger, Yu, Renz, and
  Geiger]{transfuser}
Kashyap Chitta, Aditya Prakash, Bernhard Jaeger, Zehao Yu, Katrin Renz, and
  Andreas Geiger.
\newblock {TransFuser}: Imitation with transformer-based sensor fusion for
  autonomous driving.
\newblock \emph{IEEE Transactions on Pattern Analysis and Machine Intelligence
  (TPAMI)}, 2023.

\bibitem[Dauner et~al.(2024)Dauner, Hallgarten, Li, Weng, Huang, Yang, Li,
  Gilitschenski, Ivanovic, Pavone, Geiger, and Chitta]{navsim}
Daniel Dauner, Marcel Hallgarten, Tianyu Li, Xinshuo Weng, Zhiyu Huang, Zetong
  Yang, Hongyang Li, Igor Gilitschenski, Boris Ivanovic, Marco Pavone, Andreas
  Geiger, and Kashyap Chitta.
\newblock {NAVSIM}: Data-driven non-reactive autonomous vehicle simulation and
  benchmarking.
\newblock In \emph{Advances in Neural Information Processing Systems (NeurIPS)
  Datasets and Benchmarks Track}, 2024.

\bibitem[Dosovitskiy et~al.(2017)Dosovitskiy, Ros, Codevilla, L\'opez, and
  Koltun]{carla}
Alexey Dosovitskiy, German Ros, Felipe Codevilla, Antonio L\'opez, and Vladlen
  Koltun.
\newblock {CARLA}: An open urban driving simulator.
\newblock In \emph{Conference on Robot Learning (CoRL)}, 2017.

\bibitem[Guo et~al.(2025)Guo, Liu, Wu, Pan, and Lv]{ipad}
Ke Guo, Haochen Liu, Xiaojun Wu, Jia Pan, and Chen Lv.
\newblock ipad: Iterative proposal-centric {End-to-End} autonomous driving.
\newblock \emph{arXiv preprint arXiv:2505.15111}, 2025.

\bibitem[Guo et~al.(2026)Guo, Yang, Han, Zhang, Li, Wan, and Zhang]{flowad}
Mingzhe Guo, Yixiang Yang, Chuanrong Han, Rufeng Zhang, Shirui Li, Ji Wan, and
  Zhipeng Zhang.
\newblock {FlowAD}: {Ego-Scene} interactive modeling for autonomous driving.
\newblock \emph{arXiv preprint arXiv:2603.13399}, 2026.

\bibitem[Hamdan et~al.(2025)Hamdan, Sima, Yang, Li, and Güney]{eta}
Shadi Hamdan, Chonghao Sima, Zetong Yang, Hongyang Li, and Fatma Güney.
\newblock {ETA}: Efficiency through thinking ahead, a dual approach to
  {Self-Driving} with large models.
\newblock \emph{arXiv preprint arXiv:2506.07725}, 2025.

\bibitem[He et~al.(2015)He, Zhang, Ren, and Sun]{resnet}
Kaiming He, Xiangyu Zhang, Shaoqing Ren, and Jian Sun.
\newblock Deep residual learning for image recognition.
\newblock \emph{arXiv preprint arXiv:1512.03385}, 2015.

\bibitem[Hu et~al.(2022)Hu, Chen, Wu, Li, Yan, and Tao]{stp3}
Shengchao Hu, Li Chen, Penghao Wu, Hongyang Li, Junchi Yan, and Dacheng Tao.
\newblock {ST-P3}: End-to-end vision-based autonomous driving via
  spatial-temporal feature learning.
\newblock In \emph{European Conference on Computer Vision (ECCV)}, 2022.

\bibitem[Hu et~al.(2023)Hu, Yang, Chen, Li, Sima, Zhu, Chai, Du, Lin, Wang, Lu,
  Jia, Liu, Dai, Qiao, and Li]{uniad}
Yihan Hu, Jiazhi Yang, Li Chen, Keyu Li, Chonghao Sima, Xizhou Zhu, Siqi Chai,
  Senyao Du, Tianwei Lin, Wenhai Wang, Lewei Lu, Xiaosong Jia, Qiang Liu,
  Jifeng Dai, Yu Qiao, and Hongyang Li.
\newblock Planning-oriented autonomous driving.
\newblock In \emph{Proceedings of the IEEE/CVF Conference on Computer Vision
  and Pattern Recognition (CVPR)}, 2023.

\bibitem[Huang et~al.(2021)Huang, Huang, Zhu, Ye, and Du]{bevdet}
Junjie Huang, Guan Huang, Zheng Zhu, Yun Ye, and Dalong Du.
\newblock {BEVDet}: High-performance multi-camera {3D} object detection in
  bird-eye-view.
\newblock \emph{arXiv preprint arXiv:2112.11790}, 2021.

\bibitem[Jia et~al.(2023)Jia, Gao, Chen, Yan, Liu, and Li]{jia2023driveadapter}
Xiaosong Jia, Yulu Gao, Li Chen, Junchi Yan, Patrick~Langechuan Liu, and
  Hongyang Li.
\newblock {DriveAdapter}: Breaking the coupling barrier of perception and
  planning in end-to-end autonomous driving.
\newblock In \emph{Proc. {ICCV}}, 2023.

\bibitem[Jia et~al.(2024)Jia, Yang, Li, Zhang, and Yan]{bench2drive}
Xiaosong Jia, Zhenjie Yang, Qifeng Li, Zhiyuan Zhang, and Junchi Yan.
\newblock {Bench2Drive}: Towards multi-ability benchmarking of closed-loop
  end-to-end autonomous driving.
\newblock In \emph{Advances in Neural Information Processing Systems (NeurIPS)
  Datasets and Benchmarks Track}, 2024.

\bibitem[Jia et~al.(2025)Jia, You, Zhang, and Yan]{drivetransformer}
Xiaosong Jia, Junqi You, Zhiyuan Zhang, and Junchi Yan.
\newblock {DriveTransformer}: Unified transformer for scalable end-to-end
  autonomous driving.
\newblock In \emph{International Conference on Learning Representations
  (ICLR)}, 2025.

\bibitem[Jiang et~al.(2023)Jiang, Chen, Xu, Liao, Chen, Zhou, Zhang, Liu,
  Huang, and Wang]{vad}
Bo Jiang, Shaoyu Chen, Qing Xu, Bencheng Liao, Jiajie Chen, Helong Zhou, Qian
  Zhang, Wenyu Liu, Chang Huang, and Xinggang Wang.
\newblock {VAD}: Vectorized scene representation for efficient autonomous
  driving.
\newblock In \emph{Proceedings of the IEEE/CVF International Conference on
  Computer Vision (ICCV)}, 2023.

\bibitem[Li et~al.(2025{\natexlab{a}})Li, Li, Lan, Xie, Zhang, Liu, Wu, Yu, and
  Alvarez]{hydramdppp}
Kailin Li, Zhenxin Li, Shiyi Lan, Yuan Xie, Zhizhong Zhang, Jiayi Liu, Zuxuan
  Wu, Zhiding Yu, and Jose~M. Alvarez.
\newblock {Hydra-MDP}++: Advancing {End-to-End} driving via {Expert-Guided}
  {Hydra-Distillation}.
\newblock \emph{arXiv preprint arXiv:2503.12820}, 2025{\natexlab{a}}.

\bibitem[Li et~al.(2025{\natexlab{b}})Li, Fan, He, Wang, Chen, Zhang, and
  Tan]{law}
Yingyan Li, Lue Fan, Jiawei He, Yuqi Wang, Yuntao Chen, Zhaoxiang Zhang, and
  Tieniu Tan.
\newblock Enhancing end-to-end autonomous driving with latent world model.
\newblock In \emph{International Conference on Learning Representations
  (ICLR)}, 2025{\natexlab{b}}.

\bibitem[Li et~al.(2022)Li, Wang, Li, Xie, Sima, Lu, Qiao, and Dai]{bevformer}
Zhiqi Li, Wenhai Wang, Hongyang Li, Enze Xie, Chonghao Sima, Tong Lu, Yu Qiao,
  and Jifeng Dai.
\newblock {BEVFormer}: Learning bird's-eye-view representation from
  multi-camera images via spatiotemporal transformers.
\newblock In \emph{European Conference on Computer Vision (ECCV)}, 2022.

\bibitem[Li et~al.(2024{\natexlab{a}})Li, Li, Wang, Lan, Yu, Ji, Li, Zhu,
  Kautz, Wu, Jiang, and Alvarez]{hydramdp}
Zhenxin Li, Kailin Li, Shihao Wang, Shiyi Lan, Zhiding Yu, Yishen Ji, Zhiqi Li,
  Ziyue Zhu, Jan Kautz, Zuxuan Wu, Yu-Gang Jiang, and Jose~M. Alvarez.
\newblock {Hydra-MDP}: End-to-end multimodal planning with multi-target
  hydra-distillation.
\newblock \emph{arXiv preprint arXiv:2406.06978}, 2024{\natexlab{a}}.

\bibitem[Li et~al.(2024{\natexlab{b}})Li, Yu, Lan, Li, Kautz, Lu, and
  Alvarez]{bevplanner}
Zhiqi Li, Zhiding Yu, Shiyi Lan, Jiahan Li, Jan Kautz, Tong Lu, and Jose~M.
  Alvarez.
\newblock Is ego status all you need for open-loop end-to-end autonomous
  driving?
\newblock In \emph{Proceedings of the IEEE/CVF Conference on Computer Vision
  and Pattern Recognition (CVPR)}, 2024{\natexlab{b}}.

\bibitem[Li et~al.(2025{\natexlab{c}})Li, Wang, Lan, Yu, Wu, and
  Alvarez]{hydranext}
Zhenxin Li, Shihao Wang, Shiyi Lan, Zhiding Yu, Zuxuan Wu, and Jose~M. Alvarez.
\newblock {Hydra-NeXt}: Robust {Closed-Loop} driving with {Open-Loop} training.
\newblock \emph{arXiv preprint arXiv:2503.12030}, 2025{\natexlab{c}}.

\bibitem[Li et~al.(2025{\natexlab{d}})Li, Yao, Wang, Sun, Chen, Chang, Shen,
  Wu, Lan, and Alvarez]{gtrs}
Zhenxin Li, Wenhao Yao, Zi Wang, Xinglong Sun, Joshua Chen, Nadine Chang,
  Maying Shen, Zuxuan Wu, Shiyi Lan, and Jose~M. Alvarez.
\newblock Generalized trajectory scoring for end-to-end multimodal planning.
\newblock \emph{arXiv preprint arXiv:2506.06664}, 2025{\natexlab{d}}.

\bibitem[Liao et~al.(2025)Liao, Chen, Yin, Jiang, Wang, Yan, Zhang, Li, Zhang,
  Zhang, and Wang]{diffusiondrive}
Bencheng Liao, Shaoyu Chen, Haoran Yin, Bo Jiang, Cheng Wang, Sixu Yan, Xinbang
  Zhang, Xiangyu Li, Ying Zhang, Qian Zhang, and Xinggang Wang.
\newblock {DiffusionDrive}: Truncated diffusion model for end-to-end autonomous
  driving.
\newblock In \emph{Proceedings of the IEEE/CVF Conference on Computer Vision
  and Pattern Recognition (CVPR)}, 2025.

\bibitem[Lin et~al.(2022)Lin, Lin, Pei, Huang, and Su]{sparse4d}
Xuewu Lin, Tianwei Lin, Zixiang Pei, Lichao Huang, and Zhizhong Su.
\newblock {Sparse4D}: Multi-view {3D} object detection with sparse
  spatial-temporal fusion.
\newblock \emph{arXiv preprint arXiv:2211.10581}, 2022.

\bibitem[Liu et~al.(2023)Liu, Teng, Lu, Wang, and Wang]{sparsebev}
Haisong Liu, Yao Teng, Tao Lu, Haiguang Wang, and Limin Wang.
\newblock {SparseBEV}: {High-Performance} sparse 3d object detection from
  {Multi-Camera} videos.
\newblock \emph{arXiv preprint arXiv:2308.09244}, 2023.

\bibitem[Liu et~al.(2022)Liu, Wang, Zhang, and Sun]{petr}
Yingfei Liu, Tiancai Wang, Xiangyu Zhang, and Jian Sun.
\newblock {PETR}: Position embedding transformation for multi-view {3D} object
  detection.
\newblock In \emph{European Conference on Computer Vision (ECCV)}, 2022.

\bibitem[Philion and Fidler(2020)]{lss}
Jonah Philion and Sanja Fidler.
\newblock Lift, splat, shoot: Encoding images from arbitrary camera rigs by
  implicitly unprojecting to {3D}.
\newblock In \emph{European Conference on Computer Vision (ECCV)}, 2020.

\bibitem[Renz et~al.(2025)Renz, Chen, Arani, and Sinavski]{renz2025simlingo}
Katrin Renz, Long Chen, Elahe Arani, and Oleg Sinavski.
\newblock {SimLingo}: Vision-only closed-loop autonomous driving with
  language-action alignment.
\newblock In \emph{Proc. {CVPR}}, 2025.

\bibitem[Song et~al.(2025)Song, Jia, Liu, Pan, Zhang, Wang, Zhang, Xu, Yang,
  and Luo]{momad}
Ziying Song, Caiyan Jia, Lin Liu, Hongyu Pan, Yongchang Zhang, Junming Wang,
  Xingyu Zhang, Shaoqing Xu, Lei Yang, and Yadan Luo.
\newblock Don't shake the wheel: {Momentum-Aware} planning in {End-to-End}
  autonomous driving.
\newblock \emph{arXiv preprint arXiv:2503.03125}, 2025.

\bibitem[Su et~al.(2024)Su, Wu, Yang, and Guan]{difsd}
Haisheng Su, Wei Wu, Zhenjie Yang, and Isabel Guan.
\newblock {EgoFSD}: {Ego-Centric} fully sparse paradigm with uncertainty
  denoising and iterative refinement for efficient {End-to-End} {Self-Driving}.
\newblock \emph{arXiv preprint arXiv:2409.09777}, 2024.

\bibitem[Sun et~al.(2025)Sun, Lin, Shi, Zhang, Wu, and Zheng]{sparsedrive}
Wenchao Sun, Xuewu Lin, Yining Shi, Chuang Zhang, Haoran Wu, and Sifa Zheng.
\newblock {SparseDrive}: End-to-end autonomous driving via sparse scene
  representation.
\newblock In \emph{Proceedings of the IEEE International Conference on Robotics
  and Automation (ICRA)}, 2025.

\bibitem[Tang et~al.(2025)Tang, Xu, Meng, and Cheng]{hipad}
Yingqi Tang, Zhuoran Xu, Zhaotie Meng, and Erkang Cheng.
\newblock {HiP-AD}: Hierarchical and {Multi-Granularity} planning with
  deformable attention for autonomous driving in a single decoder.
\newblock \emph{arXiv preprint arXiv:2503.08612}, 2025.

\bibitem[Wang et~al.(2021)Wang, Guizilini, Zhang, Wang, Zhao, and
  Solomon]{detr3d}
Yue Wang, Vitor Guizilini, Tianyuan Zhang, Yilun Wang, Hang Zhao, and Justin
  Solomon.
\newblock {DETR3D}: {3D} object detection from multi-view images via
  {3D}-to-{2D} queries.
\newblock In \emph{Conference on Robot Learning (CoRL)}, 2021.

\bibitem[Weng et~al.(2024)Weng, Ivanovic, Wang, Wang, and Pavone]{paradrive}
Xinshuo Weng, Boris Ivanovic, Yan Wang, Yue Wang, and Marco Pavone.
\newblock {PARA-Drive}: Parallelized architecture for real-time autonomous
  driving.
\newblock In \emph{Proceedings of the IEEE/CVF Conference on Computer Vision
  and Pattern Recognition (CVPR)}, 2024.

\bibitem[Wozniak et~al.(2025)Wozniak, Liu, Cai, and Jensfelt]{prix}
Maciej~K. Wozniak, Lianhang Liu, Yixi Cai, and Patric Jensfelt.
\newblock {PRIX}: Learning to plan from raw pixels for {End-to-End} autonomous
  driving.
\newblock \emph{arXiv preprint arXiv:2507.17596}, 2025.

\bibitem[Wu et~al.(2022)Wu, Jia, Chen, Yan, Li, and Qiao]{wu2022tcp}
Penghao Wu, Xiaosong Jia, Li Chen, Junchi Yan, Hongyang Li, and Yu Qiao.
\newblock Trajectory-guided control prediction for end-to-end autonomous
  driving: A simple yet strong baseline.
\newblock In \emph{Advances in Neural Information Processing Systems}, 2022.

\bibitem[Yao et~al.(2025)Yao, Li, Lan, Wang, Sun, Alvarez, and Wu]{drivesuprim}
Wenhao Yao, Zhenxin Li, Shiyi Lan, Zi Wang, Xinglong Sun, Jose~M. Alvarez, and
  Zuxuan Wu.
\newblock {DriveSuprim}: Towards precise trajectory selection for {End-to-End}
  planning.
\newblock \emph{arXiv preprint arXiv:2506.06659}, 2025.

\bibitem[Zhai et~al.(2023)Zhai, Feng, Du, Mao, Liu, Tan, Zhang, Ye, and
  Wang]{admlp}
Jiang-Tian Zhai, Ze Feng, Jihao Du, Yongqiang Mao, Jiang-Jiang Liu, Zichang
  Tan, Yifu Zhang, Xiaoqing Ye, and Jingdong Wang.
\newblock Rethinking the open-loop evaluation of end-to-end autonomous driving
  in {nuScenes}.
\newblock \emph{arXiv preprint arXiv:2305.10430}, 2023.

\bibitem[Zhang et~al.(2025)Zhang, Song, Jin, and Zhang]{bridgead}
Bozhou Zhang, Nan Song, Xin Jin, and Li Zhang.
\newblock Bridging past and future: {End-to-End} autonomous driving with
  historical prediction and planning.
\newblock \emph{arXiv preprint arXiv:2503.14182}, 2025.

\bibitem[Zhang et~al.(2024)Zhang, Wang, Zhu, Zhao, Chen, Zhang, Gong, Zhou,
  Zhang, Wang, Tan, Zhou, Xu, Yao, Zhang, Liu, Di, and Li]{sparsead}
Diankun Zhang, Guoan Wang, Runwen Zhu, Jianbo Zhao, Xiwu Chen, Siyu Zhang,
  Jiahao Gong, Qibin Zhou, Wenyuan Zhang, Ningzi Wang, Feiyang Tan, Hangning
  Zhou, Ziyao Xu, Haotian Yao, Chi Zhang, Xiaojun Liu, Xiaoguang Di, and Bin
  Li.
\newblock {SparseAD}: Sparse {Query-Centric} paradigm for efficient
  {End-to-End} autonomous driving.
\newblock \emph{arXiv preprint arXiv:2404.06892}, 2024.

\bibitem[Zhang et~al.(2026)Zhang, Fu, Xu, Dai, Liu, and Wang]{resworld}
Jinqing Zhang, Zehua Fu, Zelin Xu, Wenying Dai, Qingjie Liu, and Yunhong Wang.
\newblock {ResWorld}: Temporal residual world model for {End-to-End} autonomous
  driving.
\newblock \emph{arXiv preprint arXiv:2602.10884}, 2026.

\bibitem[Zheng et~al.(2024)Zheng, Song, Guo, Zhang, and Chen]{genad}
Wenzhao Zheng, Ruiqi Song, Xianda Guo, Chenming Zhang, and Long Chen.
\newblock {GenAD}: Generative end-to-end autonomous driving.
\newblock In \emph{European Conference on Computer Vision (ECCV)}, 2024.

\bibitem[Zhu et~al.(2021)Zhu, Su, Lu, Li, Wang, and Dai]{zhu2021deformable}
Xizhou Zhu, Weijie Su, Lewei Lu, Bin Li, Xiaogang Wang, and Jifeng Dai.
\newblock {Deformable DETR}: Deformable transformers for end-to-end object
  detection.
\newblock In \emph{International Conference on Learning Representations
  (ICLR)}, 2021.

\end{thebibliography}
}
\end{document}